\documentclass{article}
\usepackage{times}

\usepackage{multicol}
\usepackage[dvipsnames]{xcolor}
\usepackage{graphicx}
\usepackage{hyperref}
\usepackage{amsmath}
\usepackage{cleveref}
\usepackage[ruled,vlined,linesnumbered]{algorithm2e}
\usepackage[font=footnotesize,labelfont=bf]{caption}
\usepackage{subcaption}
\usepackage{bbm}
\usepackage{float}
\usepackage{color, soul} 
\usepackage{wrapfig,lipsum,booktabs}

\usepackage{array}
\usepackage[utf8]{inputenc}
\usepackage{amsmath, amssymb}
\usepackage{booktabs}
\usepackage{dsfont}
\usepackage{diagbox}
\usepackage{multirow}
\usepackage[preprint]{corl_2026} 
\usepackage[table]{xcolor}
\usepackage{array}
\usepackage{subcaption}

\title{MDF++: Visuo-Tactile World Action Model with End-to-end Representation Learning}

\author{
  Zixuan Huang \quad Jinhao Yu \quad
  Zachary Charlick \quad Dmitry Berenson\\
  Department of Robotics\\
  University of Michigan, Ann Arbor, United States\\
  \texttt{\{zixuanh,rickyu,zsc,dmitryb\}@umich.edu}
}

\begin{document}
\maketitle
\newcommand{\method}{\textsc{MDF++}}

\vspace{-5mm}
\begin{abstract}
	Recently, world-action models (WAMs) have emerged as a powerful framework for robot manipulation. By jointly modeling action generation and future observation prediction, WAMs can act and anticipate the outcomes of the actions. However, existing WAMs typically rely on frozen pretrained encoders trained with reconstruction, which may not produce representations well suited for robot control. 
    In this paper, we tackle the problem of learning physically-grounded, control-relevant representations for WAMs. We propose MDF++, a visuo-tactile world-action model that learns latent representations of multisensory observations end-to-end with a diffusion backbone. While naive joint training can lead to feature collapse, MDF++ introduces a multi-task learning strategy that leverages cross-modal correlations to learn task-relevant representations. We evaluate MDF++ on a suite of contact-rich manipulation tasks, including real-world bolt screwing and nut mating, and show that it outperforms WAMs built on frozen pretrained representations by up to $50\%$. Additional videos are available on our website:
\href{https://jointmdf.github.io/}{\texttt{jointmdf.github.io}}.

\end{abstract}

\begin{figure}[h]
    \centering
    \includegraphics[width=1\linewidth]{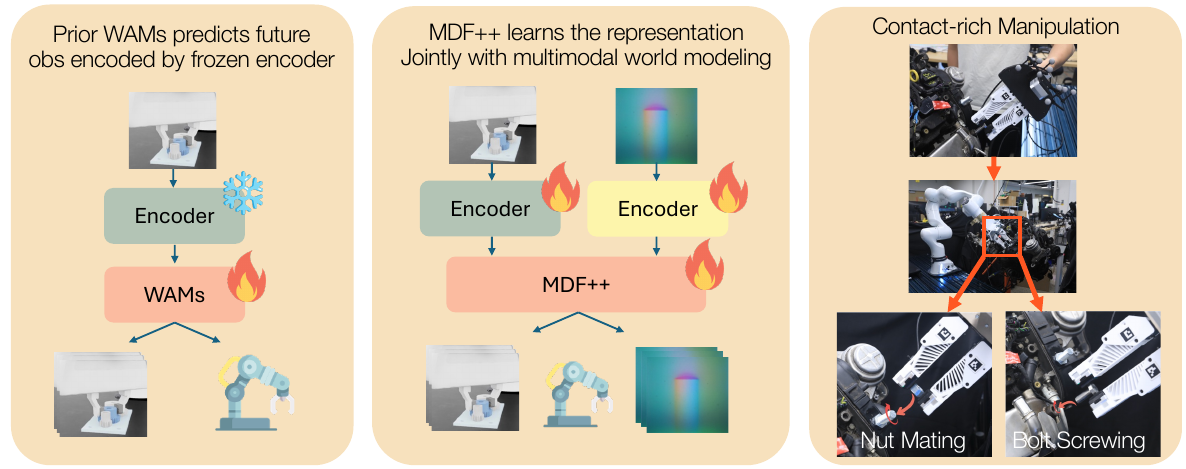}
    \caption{Existing WAMs typically predict future observations encoded by a frozen pretrained encoder optimized for visual generation or semantic understanding, which maybe suboptimal for robot manipulation. We propose MDF++, a visuo-tactile world-action model that learns task-aware representations end-to-end with cross-modality prediction. }
    \label{fig:placeholder}
\end{figure}


\section{Introduction}

When tightening a screw or mating a nut, vision alone is often insufficient: successful execution depends on subtle contact cues, force feedback, and proprioceptive information. In order to perform such tasks, it would also be useful to 
anticipate how the robot's actions affect the world, such as how contact will form, how an object will move, and how force will propagate. This motivates robot models that jointly reason about actions and their consequences. World-action models (WAMs)~\citep{cheang2024gr,li2025unified,won2025dual,wu2024unleashing,hu2024video,kim2026cosmos,liang2025video,pai2025mimic,zhu2025unified, huang2025unified} address this by jointly modeling future observations and actions within a single generative model. Recent work~\citep{li2025unified,zhu2025unified,huang2025unified} shows that such auxiliary dynamics learning can improve model capability and generalization in robot control.


However, existing WAMs still have two key limitations for robot manipulation. First, most prior WAMs are built primarily around visual observations, leaving tactile and force information largely unexplored. Second, these models typically rely on frozen pretrained encoders, often trained with pixel reconstruction or self-supervised visual objectives. While such encoders provide useful features for image generation or semantic understanding, their representations are not necessarily aligned with robot control. This issue is even more pronounced for tactile, where broadly-applicable pretrained encoders are less established. A natural solution is to finetune the encoders end-to-end with the WAM objective, but we find that naive finetuning leads to feature collapse: the encoder and diffusion backbone co-adapt to produce latents that are easy to denoise but less useful for control.

To address the above limitations, we present MDF++, a visuo-tactile world-action model that learns task-relevant representations end-to-end with a latent diffusion backbone. We build on Multimodal Diffusion Forcing (MDF)~\citep{huang2025unified}, but instead of modeling dynamics in a frozen latent space, MDF++ finetunes observation encoders jointly with the world-action model under a mixture of task-oriented objectives, including policy learning, joint action--observation prediction, forward dynamics modeling, and inverse dynamics modeling. To avoid feature collapse during joint training, MDF++ introduces a cross-modal training strategy with selective gradient back-propagation, which encourages the latent space to capture the interplay between different modalities, yielding representations that are better aligned with robot control.

Our key contributions are three-fold. First, we propose MDF++, a diffusion-based visuo-tactile WAM that learns task-relevant multimodal representations end-to-end using cross-modal prediction. Second, we introduce a selective gradient-flow strategy that enables stable joint training of observation encoders and the world-action model while avoiding feature collapse. Third, we evaluate MDF++ on seven contact-rich manipulation tasks, including two real-world threading tasks requiring millimeter-level precision. In simulation, MDF++ outperforms frozen-encoder baselines by 28$\%$ and naive joint training by 22$\%$. In real, it outperforms video-based WAMs~\citep{zhu2025unified} by 29$\%$.

\vspace{-2mm}
\section{Related Work}
\subsection{World modeling for Robotics}
Recently, world models have shown great promise for robotics. One line of work studies action-conditioned world models (WAs) to predict the future outcomes of actions, enabling data generation~\citep{team2025gigabrain, wang2026interactive}, planning~\citep{du2024video,zhou2024dino,maes2026leworldmodel,assran2025v}, and policy evaluation~\citep{team2025evaluating, wang2026interactive, zhang2025real}.
Another line of work, world action models (WAMs), jointly predicts future observations and actions~\citep{cheang2024gr,li2025unified,won2025dual,wu2024unleashing,hu2024video,kim2026cosmos,liang2025video,pai2025mimic,zhu2025unified, huang2025unified, zheng2026omnivta}. However, most prior work is limited to vision, whereas MDF++ captures the dynamics of multimodal sequences including tactile, force and robot proprioception.

Beyond the choice of modalities, a key design question is how to obtain the representation space in which world-action dynamics are learned. Most existing WA and WAMs rely on frozen pretrained encoders, typically trained with pixel reconstruction~\citep{podell2024sdxl} or self-supervised learning objectives~\citep{oquab2023dinov2, assran2023self}.
For example, UWM~\citep{zhu2025unified} and UVA~\citep{li2025unified} use pretrained image VAEs~\citep{rombach2022high,podell2024sdxl}, Dino-WM~\citep{zhou2024dino} predicts future states in the DINO feature space~\citep{oquab2023dinov2}, and MDF~\citep{huang2025unified} builds on a diffusion-based point-cloud VAE~\citep{luo2021diffusion}.
While these encoders provide useful features for visual generation or semantic understanding, they are not necessarily aligned with robot control. As a result, the world model may learn dynamics in a representation space that is suboptimal for manipulation. In contrast, MDF++ starts from a pretrained encoder and finetunes it end-to-end with the latent world-action model backbone. Through cross-modal supervision from tactile, force, and actions, MDF++ learns a representation space that is better suited for robot manipulation. FLARE~\citep{zheng2025flare} finetunes the encoder using an EMA target, but it is not designed for unified WAMs with multiple query distributions and also does not support multimodal reasoning.

\subsection{Multimodal Representation learning for Manipulation}

The integration of multi-sensory information, such as audio~\citep{du2022play, li2022see,liu2024maniwav}, force~\citep{noseworthy2025forge, kang2025robotic, wu2024tacdiffusion}, is critical for robot control. Particularly, there is growing interest in combining vision and tactile for adaptive and precise manipulation. Much of the prior works learn a visuo-tactile policy using either imitation learning~\citep{huang20243d,george2025vital, liu2025vitamin, xu2025unit, zhu2026touch} or reinforcement learning~\citep{van2016stable, suresh2024neuralfeels, lee2019making, chen2022visuo, hansen2022visuotactile, yuan2024robot, sferrazza2024power}, where representations are primarily supervised by action prediction or task reward. Other works study self-supervised tactile~\citep{cao2023learn, higuera2024sparsh, zhao2024transferable,xu2025unit} or visuo-tactile representation learning~\citep{dave2024multimodal,zhu2026touch}. In contrast, our work studies representation learning for multimodal world-action models. While these models provide richer supervision through action and cross-modal prediction, naively training encoders with diffusion losses can cause feature collapse. We address this with selective gradient flow and show that the proposed method outperforms policy-only training and pretrained self-supervised encoders~\citep{higuera2024sparsh}.


\vspace{-1mm}
\section{Preliminaries}
\label{sec:preliminaries}

\subsection{Latent Diffusion Models}
\label{sec:diffusion_models}

Diffusion models learn a data distribution by gradually corrupting clean samples with Gaussian noise and training a neural network to reverse this process~\cite{ho2020denoising}. Given a clean sample $\mathbf{x}_0$, the forward process produces a noisy sample $\mathbf{x}_k = \sqrt{\bar{\alpha}_k}\mathbf{x}_0 + \sqrt{1-\bar{\alpha}_k}\boldsymbol{\epsilon}$ at diffusion step $k$, where $\boldsymbol{\epsilon}\sim\mathcal{N}(\mathbf{0},\mathbf{I})$ and $\bar{\alpha}_k$ is determined by a noise schedule. The reverse model is trained to denoise $\mathbf{x}_k$ by predicting the injected noise, yielding the standard objective $\mathbb{E}_{k,\mathbf{x}_0,\boldsymbol{\epsilon}}\left[\left\|\boldsymbol{\epsilon}-\boldsymbol{\epsilon}_{\theta}(\mathbf{x}_k,k)\right\|^2\right]$. At inference time, samples are generated by iterative denoising from Gaussian noise. Latent diffusion models perform this denoising process in a learned latent space rather than directly in the input space~\cite{rombach2022high}. A sample $\mathbf{x}_0$ is first encoded into a latent representation $\mathbf{z}_0=\mathcal{E}(\mathbf{x}_0)$, where the encoder $\mathcal{E}$ is typically pretrained and kept frozen during diffusion training.

\subsection{Unified World-Action Model}
\label{sec:prelin:unified_world_action_model}

A unified world-action model (WAM) is a single generative model over observations and actions that can be queried under different conditioning patterns. Let $\mathbf{o}_{i:j}$ and $\mathbf{a}_{i:j}$ denote observation and action sequences from timestep $i$ to $j$, respectively. Given an observation history $\mathbf{o}_{t-h:t}$ and a prediction horizon $H$, the same model can be used as a policy, a forward dynamics model, an inverse dynamics model, or a joint action--observation predictor by choosing which variables are conditioned on and which variables are generated: \textbf{policy} $p(\mathbf{a}_{t+1:t+H} \mid \mathbf{o}_{t-h:t})$, \textbf{world-action model} $p(\mathbf{o}_{t+1:t+H},\, \mathbf{a}_{t+1:t+H} \mid \mathbf{o}_{t-h:t})$, \textbf{world model / forward dynamics} $p(\mathbf{o}_{t+1:t+H} \mid \mathbf{o}_{t-h:t},\, \mathbf{a}_{t+1:t+H})$, and \textbf{inverse dynamics model} $p(\mathbf{a}_{t+1:t+H} \mid \mathbf{o}_{t-h:t+H})$.

\begin{figure*}[!t]
    \centering
    \includegraphics[width=1\linewidth]{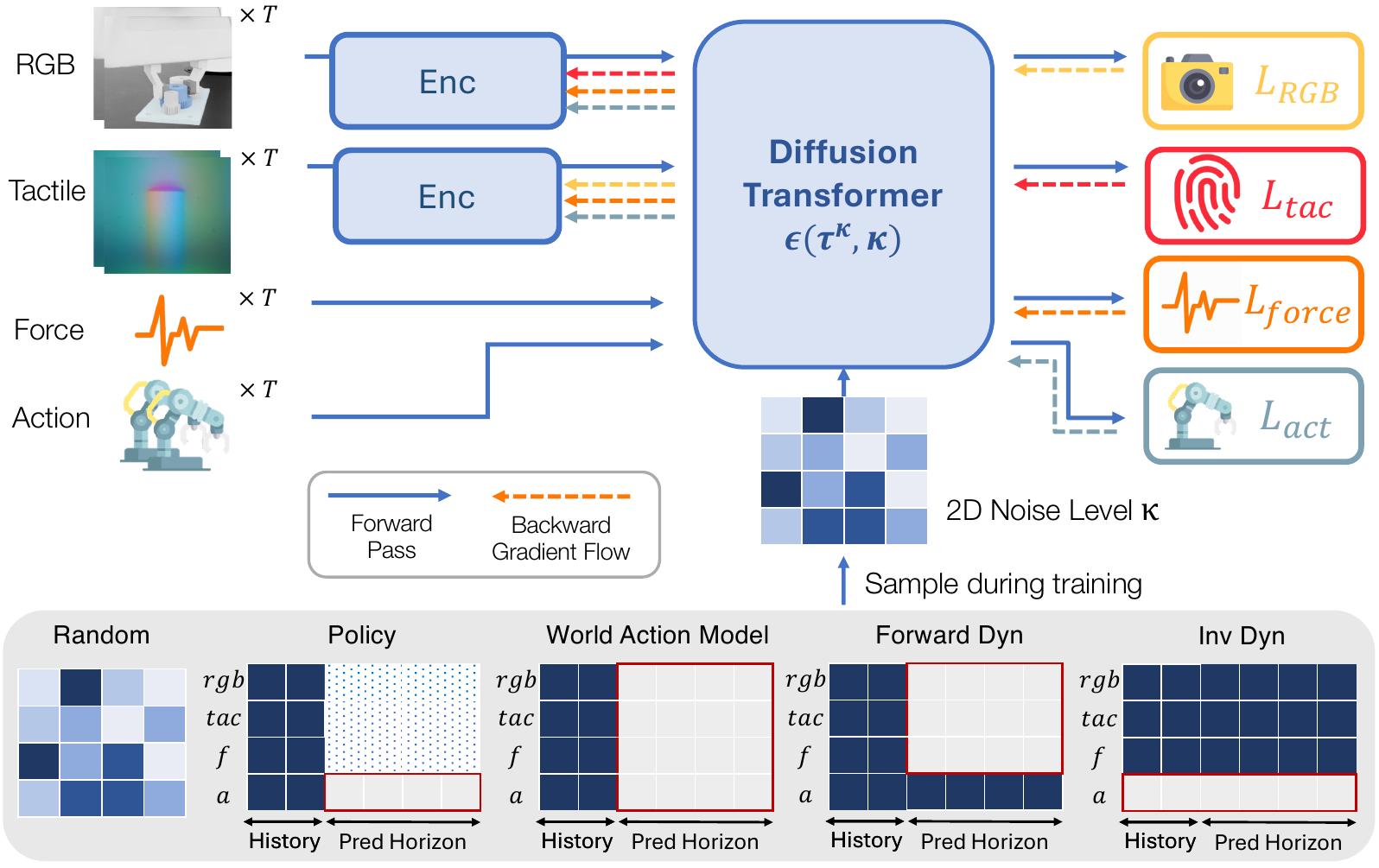}
    \vspace{-4mm}
    \caption{\textbf{MDF++ training overview.}
MDF++ jointly trains observation encoders with a latent diffusion world-action model. During training, it samples structured 2D noise patterns for control-relevant tasks, including policy learning, world-action modeling, forward/inverse dynamics, and random masking, to capture temporal and cross-modal dependencies. To avoid collapsed representations that are easy to denoise but weak for control, MDF++ blocks each modality’s denoising loss from updating its own encoder. Instead, each encoder is trained through cross-modal prediction losses, encouraging task-relevant cues for predicting other sensory modalities and actions.
    }
    \label{fig:system_fig}
    \vspace{-5mm}
\end{figure*}

\subsection{Problem Formulation}
\label{sec:problem_formulation}
We consider diffusion-based world-action modeling for contact-rich robot manipulation from an offline dataset of expert demonstrations. At each timestep $t$, the robot receives a multimodal observation $\mathbf{o}_t = \{\mathbf{o}_t^m\}_{m\in\mathcal{O}}$, where $\mathcal{O}$ denotes the set of observation modalities, such as vision, tactile sensing, force/torque sensing, and proprioception. The robot action is denoted by $\mathbf{a}_t$.

High-dimensional observations such as image and tactile are mapped into a latent representation by a modality-specific encoder $\phi_m$: $\mathbf{z}_t^m = \phi_m(\mathbf{o}_t^m)$, where $m\in\mathcal{O}$. Together with the action stream, these latent observation tokens define a multimodal trajectory
\[
    \mathbf{x}_{1:T}
    =
    \left\{
        \left(\{\mathbf{z}_t^m\}_{m\in\mathcal{O}},\, \mathbf{a}_t\right)
    \right\}_{t=1}^{T}.
\]

We consider the problem of learning a unified world-action model from multimodal demonstration trajectories for contact-rich manipulation. Given observations from multiple sensory modalities, our goal is to train observation encoders jointly with a diffusion-based world-action model to capture task-relevant feature. At test time, we evaluate the model primarily through its ability to generate actions for robot manipulation tasks.

\vspace{-3mm}
\section{Joint Representation Learning for Unified World-Action Model}

\label{sec:method}
\subsection{Multi-task Training with Multimodal Diffusion Forcing}
\label{method:multi_task_training}
To train a single backbone that supports the diverse conditional queries of a unified WAM (Sec.~\ref{sec:prelin:unified_world_action_model}), we adopt \emph{Multimodal Diffusion Forcing} (MDF)~\cite{huang2025unified} as our model architecture. Standard diffusion models apply a single scalar noise level $k$ to the entire sample~\cite{ho2020denoising}, imposing a fixed separation between conditioning inputs and prediction targets. This is restrictive for unified world-action modeling, where one model must support multiple prediction patterns.

MDF addresses this by treating noise as a continuous form of masking~\cite{chen2024diffusion, huang2025unified}: zero noise corresponds to an observed token, intermediate noise to a partially corrupted token, and maximum noise to a fully masked token. By assigning noise levels independently across time and modality, MDF can express different conditional distributions over the same trajectory.
Concretely, consider a trajectory of latent multimodal tokens and actions over a horizon $T$, $\tau = \{(z_t, a_t)\}_{t=1}^{T}$, where each timestep contains $M$ modality streams (i.e., vision, tactile, force, proprioception, and action). MDF defines a 2D \emph{Time--Modality Noise Level Matrix} $\mathbf{K}\in\{0,\ldots,K\}^{T\times M}$, where each entry $k_{t,m}$ specifies the diffusion step applied to modality $m$ at time $t$. The forward process corrupts each token according to its own noise level, $\tau_{t,m}^{k_{t,m}} = \sqrt{\bar{\alpha}_{k_{t,m}}}\,\tau_{t,m}^{0} + \sqrt{1-\bar{\alpha}_{k_{t,m}}}\,\epsilon_{t,m},\quad \epsilon_{t,m}\sim\mathcal{N}(\mathbf{0},\mathbf{I}).$
A diffusion transformer $\epsilon_\theta$ conditioned on both the corrupted trajectory $\tau^{\mathbf{K}}$ and the matrix $\mathbf{K}$ is then trained to predict the per-token noise with the standard DDPM objective~\cite{ho2020denoising},
\begin{equation}
    \mathcal{L}_{\text{MDF}}(\theta) = \mathbb{E}_{\tau^{0},\mathbf{K},\boldsymbol{\epsilon}}\left[\sum_{t=1}^{T}\sum_{m=1}^{M}\bigl\|\epsilon_{t,m} - \epsilon_\theta(\tau^{\mathbf{K}},\mathbf{K})_{t,m}\bigr\|^{2}\right].
\end{equation}

While the original MDF training scheme is highly flexible, it samples the noise-level matrix $\mathbf{K}$ from a purely random distribution (bottom left of Fig.~\ref{fig:system_fig}). This covers a large family of prediction tasks, since each masking pattern defines a different conditional distribution. However, many of these conditionals are only weakly related to the inference problems needed for robot manipulation.To bias training toward control-relevant queries, we train the model on five structured conditional distributions with equal sampling probability: \textbf{policy learning}, \textbf{world-action modeling}, \textbf{forward dynamics}, \textbf{inverse dynamics}, and \textbf{random masking}. As shown in Fig.~\ref{fig:system_fig}, conditioning tokens are assigned noise level zero, while prediction tokens are assigned noise levels sampled uniformly from $[0,K]$. These structured patterns focus model capacity on conditional distributions directly useful for robot control. Although our final downstream objective is action generation for visuomotor control, prior work shows that auxiliary prediction tasks, such as forward and inverse dynamics modeling, can encourage representations to capture task-relevant features and temporal structure in demonstration trajectories~\citep{cui2024dynamo,brandfonbrener2023inverse,guo2024prediction}.

However, \textbf{existing WAMs typically cannot fully leverage this multi-task supervision for representation learning}, because they rely on frozen encoders pretrained with reconstruction objectives~\citep{cen2025worldvla,li2025unified,zhu2025unified,huang2025unified}. For example, MDF~\citep{huang2025unified} uses a frozen point-cloud encoder pretrained with a reconstruction objective. Reconstruction encourages the encoder to preserve information needed to reproduce the input, including task-irrelevant details that may not be useful for downstream control. While this design provides stable geometric features, it decouples representation learning from the control and prediction objectives used by the world-action model. In the next section, we describe how MDF++ leverages this multi-task supervision to train observation encoders end-to-end with the diffusion backbone, yielding action-aware representations for robotic manipulation.

\subsection{Cross-modal Training with Selective Gradient Flow}
\label{method:selective_grad}

Given a diffusion-based WAM, the most direct way to fine-tune the observation encoders is to backpropagate the diffusion loss through the encoder. However, this naive end-to-end training scheme usually leads to collapsed feature space. Since the same latent tokens serve as both conditioning inputs and prediction targets, gradients from a modality's own denoising loss can encourage the encoder and diffusion backbone to co-adapt toward a latent space that is easy to denoise~\cite{leng2025repa}, but weak for control. In our experiments, this direct fine-tuning strategy often degrades task performance.

MDF++ avoids feature collapse with a simple selective gradient-flow rule inspired by REPA-E~\citep{leng2025repa}: the encoder for each observation modality is updated only by diffusion losses from other streams, not by the loss of its own stream. 
Let $\mathcal{O}$ denote the set of observation modalities and
$\mathcal{M}=\mathcal{O}\cup\{a\}$ denote all predicted streams, including
actions. Let $\mathcal{L}_s$ be the MDF denoising loss for stream
$s\in\mathcal{M}$. For each observation encoder $\phi_o$, where
$o\in\mathcal{O}$, MDF++ applies the masked gradient
\begin{equation}
    \nabla_{\phi_o}^{\mathrm{MDF++}} \mathcal{L}
    =
    \sum_{s\in\mathcal{M}}
    \mathbf{1}[s\neq o]\,
    \nabla_{\phi_o} \mathcal{L}_s .
    \label{eq:selective_gradient}
\end{equation}
Equivalently, when computing the denoising loss for modality $m$, gradients are stopped at the output of its own encoder $\phi_m$, while gradients from cross-modal prediction losses are allowed to update it. 

This rule trains each encoder to produce representations that are predictive of future signals in other modalities. For example, the visual encoder is not optimized to reconstruct visual appearance itself; instead, it is optimized through losses for predicting other sensory streams and actions. Since predicting tactile feedback, interaction forces, and successful actions requires information about the object, the end-effector, and their potential contact regions, this cross-modal supervision naturally encourages the visual representation to capture manipulation-relevant cues. Because no encoder is trained to denoise its own input, the learned representations are shaped by cross-modal predictive pressure rather than modality-specific reconstruction shortcuts.

\textbf{Model Architecture.}
We instantiate $\tau$ as a sequence of modality-specific tokens. RGB and tactile images are encoded as CLS tokens from pretrained ViTs~\citep{dosovitskiy2020image}, while low-dimensional streams such as force, proprioception, and actions are projected with linear layers. At each timestep, all modality tokens are concatenated with their corresponding noise-level embeddings and processed by a diffusion transformer.
\textbf{Training Recipe}. We identify several design choices that are important for training stability and performance. First, for each sampled trajectory, we sample multiple training modes to encourage robust learning across different conditioning and prediction settings and improve training efficiency. Second, MDF++ injects different noise levels into latent tokens during diffusion training, we detach the encoder outputs for highly corrupted tokens; gradients from these samples are often weakly correlated with the original observations and can destabilize representation learning. Third, we apply a cosine learning-rate schedule to the encoder and cool it down before the diffusion backbone.

\vspace{-2mm}
\section{Experiments}
\label{sec:result}
We evaluate MDF++ on a suite of simulated and real-world visuo-tactile manipulation tasks. Our experiments are designed to answer three questions: (1) Does the proposed selective gradient-flow strategy stabilize joint representation learning? (2) Are the learned task-aware representations more effective than pretrained semantic or reconstruction-based encoders for contact-rich manipulation? (3) Does multi-task training provides a better representation for contact-rich robot manipulation?

\subsection{Simulation Experiments}
\textbf{Tasks:} We consider five manipulation tasks that requires multimodal reasoning. Three precision-focused tasks are selected from Isaac Lab \textsc{Factory}~\citep{mittal2025isaaclab}: PegInsert, GearMesh, and NutThread. We adapt the simulation environments from Forge~\citep{noseworthy2025forge} by adding a GelSight Mini tactile sensor. For each task, we collect 1,000 demonstrations using an RL expert policy.
We additionally include two tactile-exploration tasks from \textsc{ManiFeel}~\citep{luu2025manifeel}. In ObjectSearch, the robot must locate, grasp, and lift a cube-shaped object randomly placed inside a closed container. In BallSorting, the robot must sort two visually similar balls using tactile sensing.

\textbf{Baselines and ablations}
We additionally compare against \textbf{MDF (adapted)}~\citep{huang2025unified}, which uses frozen observation encoders and random masking in the same visuo-tactile setting, and \textbf{MDF + selective E2E}, which retains random masking but jointly trains the encoders with selective gradient flow. Their Forge runs use end-of-trajectory padding, unlike the MDF++ reference, so the comparison does not fully isolate the objectives.
We compare MDF++ against several variants and representative baselines. \textbf{MDF++ (naive grad)} jointly trains the encoders and diffusion model by backpropagating denoising losses from all modalities through all encoders. \textbf{MDF++ (SIGReg)} adds SIGReg~\citep{balestriero2025lejepa, maes2026leworldmodel} as a latent-space regularizer to mitigate representation collapse. SIGReg optimizes a normality test over random one-dimensional projections, encouraging the latent embeddings to match an isotropic Gaussian distribution. \textbf{MDF++ (frozen-ViT)} and \textbf{MDF++ (frozen-Sparsh)} use frozen pretrained encoders for visual and tactile observations. The pretrained ViT~\citep{steiner2021train} provides a semantic representation, while Sparsh~\citep{higuera2024sparsh} is pretrained specifically for vision-based tactile sensing. These baselines test whether fixed pretrained representations are sufficient for visuo-tactile control. \textbf{UWM}~\citep{zhu2025unified} is a unified world-action model that encodes future observations using a frozen SDXL VAE~\citep{podell2024sdxl}, which is optimized for high-quality pixel reconstruction. This baseline evaluates a reconstruction-oriented latent space for world-action modeling. Finally, \textbf{DP w/ tactile} is a diffusion policy baseline that conditions on both visual and tactile observations. This baseline compares MDF++ against a policy-only model without the proposed multimodal world-action training objective.

\begin{table}[t]
\centering
\caption{\textbf{Simulation tasks and success rates on five simulated visuo-tactile manipulation tasks.} The two added MDF variants report per-task maxima across available evaluation seeds and policy/collocation modes, not seed means.}
\label{tab:main_results}

\vspace{-1mm}
\includegraphics[width=0.92\linewidth]{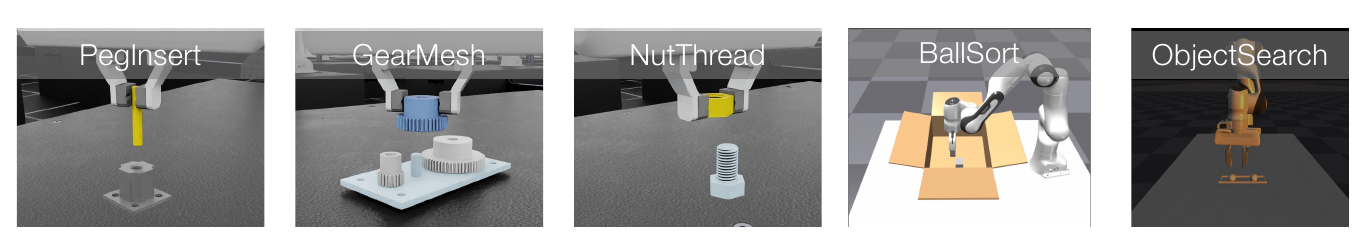}
\vspace{1mm}

\resizebox{\linewidth}{!}{
\begin{tabular}{lccccc>{\columncolor{gray!12}}c}
\toprule
\textbf{Model} 
& \textbf{PegInsert} $\uparrow$
& \textbf{GearMesh} $\uparrow$
& \textbf{NutThread} $\uparrow$
& \textbf{BallSort} $\uparrow$
& \textbf{ObjectSearch} $\uparrow$
& \textbf{Avg.} $\uparrow$ \\
\midrule
MDF++ 
& 85.3 & 78.7 & 92.0 & 93.3 & 85.3 & \textbf{86.9} \\

MDF (adapted)~\citep{huang2025unified}
& 89.3 & 78.7 & 92.7 & 42.0 & 39.3 & 68.4 \\
MDF + selective E2E
& 86.7 & 54.7 & 92.0 & 79.3 & 44.7 & 71.5 \\
\textsc{MDF++} (naive grad)
& 56.7 & 50.0 & 90.7 & 67.3 & 56.0 & 64.1 \\
MDF++(SIGReg~\citep{maes2026leworldmodel}) 
& 50.7 & 32.0 & 84.7 & 48.0 & 75.3 & 58.1 \\
MDF++(frozen-ViT) 
& 86.7 & 66.0 & 89.3 & 36.7 & 14.7 & 58.7 \\
MDF++(frozen-Sparsh) 
& 90.0 & 64.7 & 81.3 & 40.0 & 18.7 & 58.9 \\
MDF++(policy) 
& 62.7 & 64.7 & 90.7 & 90.0 & 86.6 & 78.9 \\
UWM~\citep{zhu2025unified} w/ tactile
& 88.6 & 72.0 & 89.3 & 53.3 & 68.0 & 74.2\\
DP~\cite{zhu2026touch} w/ tactile
& 88.7 & 81.3 & 91.3 & 28.0 & 54.0 & 68.7 \\
\bottomrule
\end{tabular}
}
\vspace{-5mm}
\end{table}

\subsubsection{Results}
The success rate of all methods can be found at Table.~\ref{tab:main_results}. The success rates are computed over 150 trials for each method. 

\textbf{Structured queries complement selective representation learning.}
The adapted MDF baseline achieves $68.4\%$ average success, and adding selective end-to-end representation learning under random masking increases this to $71.5\%$. Combining selective learning with structured policy, world-action, forward-dynamics, and inverse-dynamics queries yields $86.9\%$, an improvement of $18.5$ percentage points over adapted MDF. These results support coupling structured queries with joint representation learning.

\textbf{Selective gradient flow stabilizes end-to-end representation learning.}
Table~\ref{tab:main_results} shows that MDF++ achieves the best average success rate across the five simulated tasks, reaching $86.9\%$. MDF++ (naive grad), which backpropagates all denoising losses through all encoders, achieves $64.1\%$. This large gap supports our hypothesis that naive end-to-end diffusion training can produce shortcut latent representations that are easy to denoise but less useful for control. Adding SIGReg as a latent-space regularizer does not resolve this issue, achieving only $58.1\%$ average success, suggesting that the regularization is insufficient for joint encoder learning of latent diffusion models.

\textbf{Task-aware representations outperform frozen pretrained encoders.}
MDF++ substantially outperforms MDF++(frozen-ViT), MDF++(frozen-Sparsh) and UWM, which achieve average success rates of $58.7\%$, $58.9\%$ and 74.2$\%$, respectively. Although ViT and Sparsh provide strong semantic or tactile representations, their features are not optimized for the action, force, and contact dynamics required by these tasks.  UWM performs better than the frozen-encoder ablations on average, but still lags behind MDF++. This suggests that predicting future observations at the pixel level can impose unnecessary reconstruction burden, forcing the model to allocate capacity to visual details that may be irrelevant for control. This limitation is especially evident on the tactile-exploration tasks: MDF++ achieves $93.3\%$ on BallSort and $85.3\%$ on ObjectSearch, while the frozen-encoder variants perform much worse. These results indicate that contact-rich manipulation benefits from representations shaped by task-specific multimodal prediction objectives.

\textbf{World-action training improves average success over policy-only learning.}
MDF++ improves average success over MDF++(policy) from $78.9\%$ to $86.9\%$. Its largest gains are on PegInsert and GearMesh ($22.6$ and $14.0$ percentage points), with smaller gains on NutThread and BallSort ($1.3$ and $3.3$ points); ObjectSearch decreases by $1.3$ points. Compared with DP w/ tactile, MDF++ improves average success from $68.7\%$ to $86.9\%$, with large gains on BallSort and ObjectSearch ($65.3$ and $31.3$ points), although DP performs slightly better on PegInsert and GearMesh. These results suggest that auxiliary world-action objectives provide useful supervision beyond action prediction, while their benefit varies by task and baseline.


\vspace{-2mm}
\subsection{Representation Analysis}
\label{sec:representation_analysis}
We examine feature variance and effective rank for the RGB and tactile encoders, and assess action information using a linear probe trained on frozen RGB features. Lower action-prediction mean squared error (MSE) indicates that actions are more linearly predictable from the representation.

\begin{table}[t]
\centering
\caption{\textbf{Representation diagnostics.} RGB and tactile features have dimension 192. The action probe uses frozen RGB features. High variance or effective rank alone does not imply low action-prediction error.}
\label{tab:representation_analysis}
\small
\setlength{\tabcolsep}{5pt}
\begin{tabular}{lccccc}
\toprule
& \multicolumn{2}{c}{\textbf{RGB}} & \multicolumn{2}{c}{\textbf{Tactile}} & \textbf{RGB probe} \\
\cmidrule(lr){2-3}\cmidrule(lr){4-5}
\textbf{Representation} & Variance & Eff. rank & Variance & Eff. rank & MSE $\downarrow$ \\
\midrule
Pretrained ViT & 10.09 & 49.10 & 8.26 & 20.28 & 0.163 \\
Self-denoising only & 2.15 & 3.67 & 0.05 & 2.22 & 0.574 \\
Naive E2E & 7.51 & 23.51 & 2.90 & 5.99 & 0.520 \\
MDF++ & 9.62 & 29.19 & 7.75 & 33.34 & \textbf{0.043} \\
\bottomrule
\end{tabular}
\end{table}

Table~\ref{tab:representation_analysis} separates representation diversity from action predictability. The pretrained ViT has the highest RGB variance and effective rank, but its probe error is approximately $3.8\times$ that of MDF++. Self-denoising-only supervision produces low effective ranks in both modalities and a probe error of $0.574$, consistent with representation collapse. Naive end-to-end training retains more diversity, but its tactile effective rank remains low ($5.99$ versus $33.34$ for MDF++), and its probe error is $0.520$ versus $0.043$. These diagnostics support the hypothesis that selective cross-modal supervision reduces representation collapse while making action information more linearly accessible.

\subsection{Effect of Self-denoising Gradients}
\label{sec:self_denoising_ablation}
We scale each encoder's self-denoising gradients by $\lambda_{\mathrm{self}}$, retaining cross-modal gradients and backbone denoising losses. Average simulation success rates at $\lambda_{\mathrm{self}}=0,\,0.1,\,0.5,\,1$ are $86.9\%,\,80.0\%,\,74.9\%,\,64.1\%$, respectively. Zero corresponds to MDF++ and one to naive end-to-end training. Intermediate results use per-task maxima across available seeds and policy/collocation modes. The trend supports suppressing self-denoising encoder gradients, consistent with the representation diagnostics; differences in training recipes and evaluation selection limit a strictly controlled interpretation.

\subsection{Physical Experiments}
\textbf{Hardware Platforms}. We run our experiments on a Franka Research 3 robot with WSG-50 gripper. We use gelsight mini for tactile sensing and only feed the image of right finger to the policy. A Realsense 405 camera is mounted on the wrist to provide visual observation.

\textbf{Tasks} We evaluate our method on two challenging contact-rich manipulation tasks: Bolt Screwing and Nut Mating, on a real car engine. Both tasks require precise alignment under partial observability, where small pose errors can lead to failed insertion or jamming.
In Bolt Screwing, the robot must align a bolt with a threaded hole and generate the appropriate downward force and rotational motion to initiate and maintain thread engagement. In Nut Mating, the robot must align a nut with a fixed bolt and guide it into the correct threading configuration. These tasks are particularly difficult because successful execution depends on millimeter-level accuracy; even slight misalignment can prevent the threads from engaging or cause cross-threading. 

\textbf{Data Collection and Processing}

Collecting high-quality demonstrations for Bolt Screwing and Nut Mating through direct teleoperation is challenging, as both tasks require precise alignment and axis-constrained rotation~\citep{sha2026efficient}. We therefore collect demonstrations using a customized version of UMI~\citep{chi2024universal} equipped with GelSight sensors. In total, we collect 255 demonstrations for Nut Mating and 300 demonstrations for Bolt Screwing.

The original UMI formulation uses future end-effector states as actions to compensate for control delay. However, we find that this is insufficient for our tasks, since it does not account for controller tracking error. This issue is especially problematic for contact-rich manipulation, where millimeter-level accuracy is required, but blindly increasing controller gains may lead to excessive contact forces and potential hardware damage. To address this, we replay the collected trajectories in the robot frame and update the commanded actions using a residual correction proportional to the tracking error. Empirically, this reduces the average tracking error from approximately $4$ mm to $1.2$ mm, which is critical for successful execution. Please see appendix for details.

\textbf{Evaluation protocol.}
At initialization, the nut or bolt is randomly placed between the robot fingers at different regions of the GelSight sensor, and each policy is evaluated under diverse initial end-effector poses. We evaluate three policies on two real-world threading tasks with 20 trials per policy-task pair. Each rollout is scored using a stage-based protocol: successful alignment receives 60 points, successful $180^\circ$ rotation receives 75 points, and full credit is awarded only when the gripper releases and the nut or bolt remains fixed. The complete scoring rubric is provided in the appendix.

We additionally report success rates, defining success as releasing the manipulated object while it remains fixed to its threaded counterpart.

\begin{figure}[!htbp]
    \centering
    \includegraphics[width=0.45\linewidth]{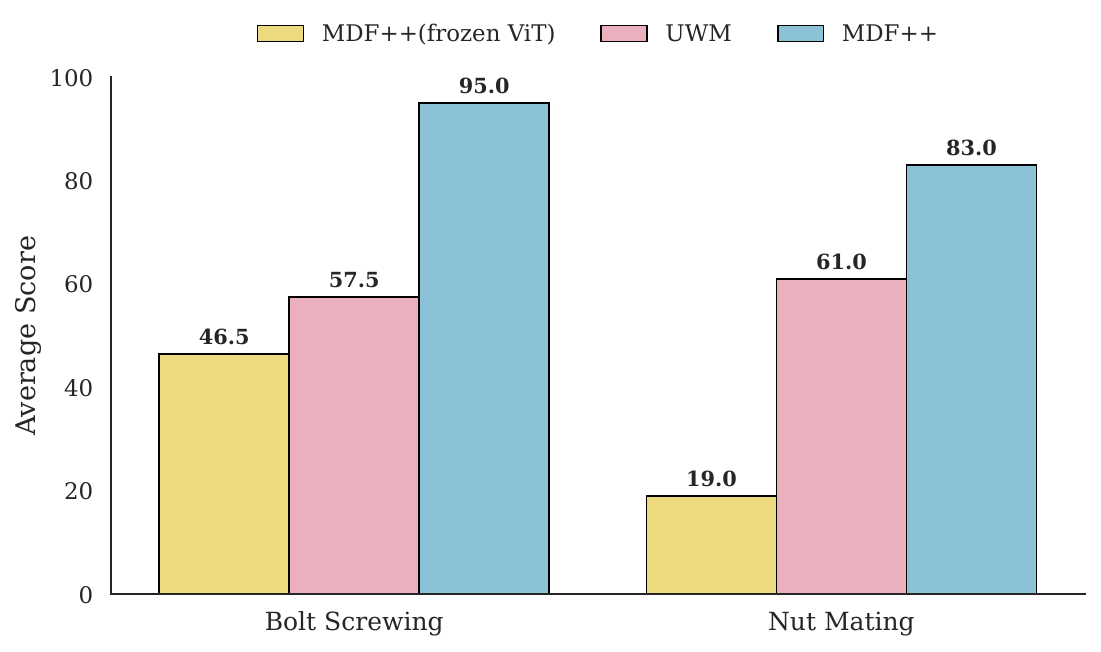}
    \caption{\textbf{Real-world scores.}
    MDF++ outperforms baselines on both threading tasks.}
    \label{fig:real_results}
\end{figure}

\textbf{Result Analysis}
Figure~\ref{fig:real_results} reports the stage-based scores for both threading tasks.
MDF++ achieves success rates of $80\%$ (16/20) on Bolt Screwing and $40\%$ (8/20) on Nut Mating, compared with $50\%$ (10/20) and $25\%$ (5/20) for UWM, and $30\%$ (6/20) and $15\%$ (3/20) for MDF++(frozen-ViT), respectively.
In our evaluations, MDF++ adapts to varied initial end-effector poses and object placements on the GelSight sensor, and recovers from small misalignments by making contact, refining alignment, and continuing rotation. UWM often fails at phase transitions, stopping before reaching the object or rotating before thread engagement. MDF++(frozen-ViT) frequently stops before contact or after coarse alignment. These observations suggest that task-aware representations help ground actions in the contact and alignment state.

\section{Conclusion}
We presented MDF++, a visuo-tactile world-action model that learns task-aware representations end-to-end with cross-modality prediction. By using multi-task world-action objectives together with selective gradient flow, MDF++ avoids pixel-level reconstruction and predicts future in a representation space that is helpful for manipulation. Experiments on simulated and real-world contact-rich manipulation tasks show that MDF++ outperforms frozen pretrained encoders, naive end-to-end finetuning, and policy-only training.
\label{sec:conclusion}

\section{Limitations and future works}
While MDF++ shows promising results for contact-rich manipulation, several limitations remain. First, MDF++ processes raw force signals at a relatively low frequency, which may miss high-frequency contact dynamics. Future work could learn force encoders over streaming high-frequency force measurements.
Second, MDF++ requires manually tuning the loss weights for different modality. Automatically balancing these objectives would reduce task-specific tuning.
Finally, although selective gradient flow helps prevent feature collapse, it also blocks gradients from each modality's own denoising loss, potentially limiting the diffusibility of the learned latent space. Future work could explore alternative latent-training strategies, such as Unified Latents~\citep{heek2026unified}, to improve diffusibility while preserving control-relevant features.

\clearpage
\acknowledgments{We thank Fan Yang and Luis Marques for helpful discussions of the ideas in this work, and Nima Fazeli for lending us a 3D printer and a WSG-50 gripper. This work was supported in part by the Office of Naval Research under grants N00014-24-1-2036 and N6833525C0329 and NSF grants IIS-2113401 and IIS-2220876.}



\appendix

\clearpage
\section*{Appendix}
\section{Implementation Details}

\subsection{\method{}}
\label{sec:mdfpp-impl}

\begin{figure}[h]
\centering
\includegraphics[width=\linewidth]{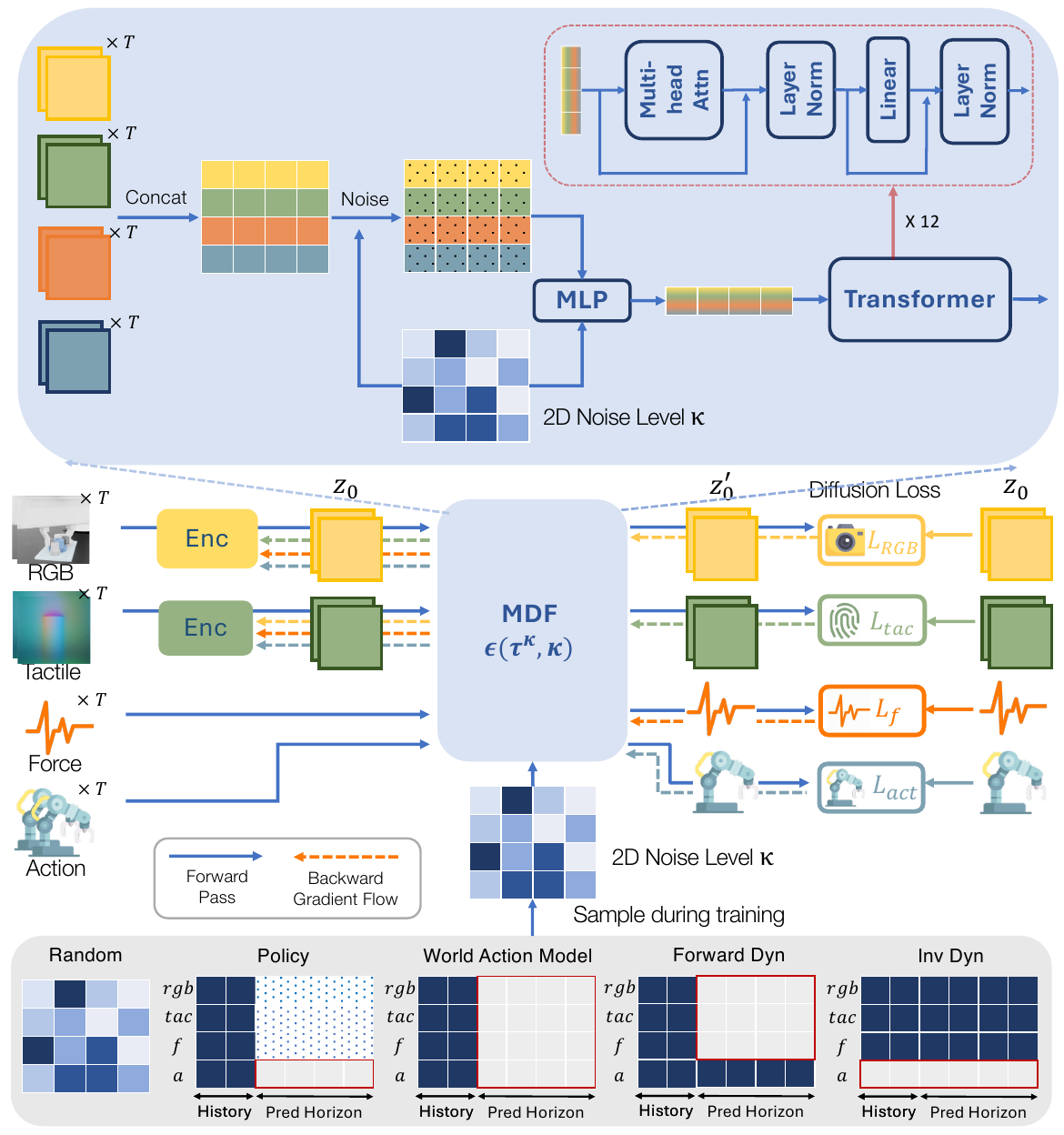}
\caption{\textbf{Detailed architecture of \method{}.} The model jointly tokenizes vision, tactile, force, proprioception, and action streams, and a diffusion transformer operates over the resulting latent tokens to predict actions and future multimodal observations.}
\label{fig:sys_figure_detailed}
\end{figure}

\paragraph{Architecture.}
Figure~\ref{fig:sys_figure_detailed} illustrates the implementation of \method{}. Each trajectory is represented as a sequence of modality-specific tokens over a history window and prediction horizon. RGB and tactile images are resized to $224 \times 224$ and encoded independently with ViT-Tiny/16 encoders, initialized from the ImageNet-21k \texttt{augreg\_in21k} checkpoint. We use the CLS token as the latent representation for each image. We use a batch norm layer to normalize the tokens. Low-dimensional streams, including force, proprioception, and actions, are mapped to latent tokens with lightweight linear projections. At each timestep, the model concatenates the available modality tokens and concatenate the tokens with diffusion-step embeddings before passing them to the diffusion transformer.

\paragraph{Diffusion and optimization.}
\method{} uses a 12-layer transformer with hidden dimension 512, 8 attention heads, and a 512-dimensional feed-forward layer. During training, we sample a 2D noise-level matrix over time and modality, which lets the same model express policy prediction, joint action--observation prediction, forward dynamics, inverse dynamics, and random masking queries. The diffusion objective predicts clean tokens ($x_0$) under a cosine noise schedule with 1000 training timesteps; DDIM sampling uses 10 steps for evaluation. We train for 200{,}000 steps with AdamW, batch size 64, bf16 mixed precision, weight decay $1\!\times\!10^{-4}$, and gradient clipping at 1.0. The diffusion backbone uses learning rate $1\!\times\!10^{-4}$, while the image encoders use a lower learning rate $1\!\times\!10^{-5}$ with 1{,}000 warm-up steps and a cosine decay that ends at 150{,}000 steps. We apply EMA to the policy weights and weight the denoising losses by modality: 4.0 for actions, 1.0 for force and proprioception, and 0.01 for RGB and tactile tokens.

\paragraph{Inference.}
At test time, the same trained model can generate actions using either the policy mode or the world-action model mode. We select the inference mode that performs better for each task.

\begin{table}[h]
\centering
\caption{Model hyperparameters for the multi-modal diffusion forcing policy.}
\label{tab:model-hparams}
\small
\begin{tabular}{lll}
\toprule
\textbf{Component} & \textbf{Hyperparameter} & \textbf{Value} \\
\midrule
\multirow{6}{*}{Diffusion transformer}
  & Architecture                & Transformer (\texttt{trans}) \\
  & Hidden dimension            & 512 \\
  & Number of layers            & 12 \\
  & Attention heads             & 8 \\
  & Feed-forward dimension      & 512 \\
  & Activation                  & ReLU \\
\midrule
\multirow{1}{*}{Conditioning embeddings}
  & Diffusion-step embedding dim ($k$) & 128 \\
\midrule
\multirow{6}{*}{Diffusion process}
  & Training timesteps           & 1000 \\
  & Sampling timesteps (DDIM)    & 10\\
  & DDIM $\eta$                  & 0.0 \\
  & Noise schedule               & cosine \\
  & Prediction objective         & $x_0$ (\texttt{pred\_x0}) \\
  & Stabilization level          & 1 \\
\midrule
\multirow{6}{*}{Image encoders (RGB / tactile)}
  & Backbone                     & ViT-Tiny/16 (\texttt{augreg\_in21k}) \\
  & Pretraining                  & ImageNet-21k \\
  & Input resolution             & $224 \times 224$ \\
  & Output token                 & CLS \\
  & Output dimension             & 192 \\
  & Output BN                    & BatchNorm ($m{=}0.1$) \\
\midrule
\multirow{2}{*}{Sequence \& I/O}
  & History length $H$           & 4 \\
  & Seq length $L$       & 10 \\
\bottomrule
\end{tabular}
\end{table}
\begin{table}[h]
\centering
\caption{Training hyperparameters.}
\label{tab:train-hparams}
\small
\begin{tabular}{ll}
\toprule
\textbf{Hyperparameter} & \textbf{Value} \\
\midrule
Optimizer                          & AdamW \\
Adam $(\beta_1, \beta_2)$          & $(0.9,\ 0.95)$ \\
Base learning rate                 & $1\!\times\!10^{-4}$ \\
Encoder learning rate              & $1\!\times\!10^{-5}$ \\
Weight decay                       & $1\!\times\!10^{-4}$ \\
Gradient clip                      & 1.0 \\
LR warm-up steps                   & 1{,}000 \\
Encoder LR cosine-decay end step   & 150{,}000 \\
Total training steps               & 200{,}000 \\
Batch size                         & 64 \\
Mixed precision                    & bf16 \\
EMA on policy weights              & enabled \\
\midrule
Conditioning sampling              & mixture (uniform over 5 modes) \\
Modality oversampling factor       & 8 \\
Modality loss weights ($\lambda$)  & action: 4.0,\ rgb/tactile: 0.01,\ force/proprio.: 1.0 \\
\bottomrule
\end{tabular}
\end{table}

\begin{table}[h]
\centering
\caption{\textbf{Input modalities and loss weights for each task family.} The custom simulated assembly tasks use force input in addition to RGB, tactile, and proprioception, whereas the tactile sorting, tactile exploration, and real-world threading tasks do not use force input. Loss weights are used when the corresponding stream is present.}
\label{tab:task-input-modalities}
\small
\begin{tabular}{@{}lccccc@{}}
\toprule
\textbf{Task family} & \textbf{RGB} & \textbf{Tactile} & \textbf{Force} & \textbf{Proprio.} & \textbf{Action} \\
\midrule
IsaacLab Factory (PegInsert, GearMesh, NutThread)
& \checkmark & \checkmark & \checkmark & \checkmark & \checkmark \\
Manifeel (BallSorting, ObjectSearch)
& \checkmark & \checkmark & -- & \checkmark & \checkmark \\
Real-world threading (Bolt Screwing, Nut Mating)
& \checkmark & \checkmark & -- & \checkmark & \checkmark \\
\midrule
Loss weight $\lambda$
& 0.01 & 0.01 & 1.0 & 1.0 & 4.0 \\
\bottomrule
\end{tabular}
\end{table}

\subsection{UWM}
\label{sec:uwm-impl}

We compare \method{} against the Unified World Model (UWM) baseline. Since the original UWM predicts future camera-view video, we adapt it to the same visuo-tactile manipulation setting used by \method{} for an apples-to-apples comparison of representation choices. The adapted UWM is trained with tactile and force observations: its conditioning input includes RGB, tactile image, force/wrench, and proprioceptive histories, and its future-observation target includes both RGB and tactile video.

The observation history is encoded by a fixed-architecture encoder into a global conditioning vector. We use a ResNet-18 image encoder pretrained on ImageNet-1k for the RGB and tactile image observations, and concatenate flattened low-dimensional tool pose and wrench observations to the encoded image feature. Future RGB and tactile frames are encoded with a frozen pretrained Stable Diffusion XL VAE (\texttt{stabilityai/sdxl-vae}), so the diffusion transformer predicts future observations in a low-resolution VAE latent space rather than directly in pixel space. These multi-view future-observation latents are patchified by a 3D convolutional patch embedding and concatenated with action tokens and learned register tokens. The diffusion transformer applies AdaLN conditioning that fuses the global observation feature with two independent diffusion timesteps---one for the action stream and one for the future-observation stream---following the dual-timestep formulation of UWM. The training loss is the sum of MSE noise-prediction losses on actions and on the future RGB/tactile VAE latents. We do not use language conditioning. The action chunk length is $7$ to match \method{}, and the observation history is $4$ frames with a total sequence length of $14$ frames. We use UWM's default DDIM scheduler with a cosine variance schedule (\texttt{squaredcos\_cap\_v2}), $100$ training steps, and $10$ inference steps. All other training settings (optimizer, batch size, mixed precision, demonstrations) follow Table~\ref{tab:train-hparams} unless noted in Table~\ref{tab:uwm-hparams}.

\begin{table}[h]
\centering
\caption{UWM baseline hyperparameters.}
\label{tab:uwm-hparams}
\small
\begin{tabular}{lll}
\toprule
\textbf{Component} & \textbf{Hyperparameter} & \textbf{Value} \\
\midrule
\multirow{6}{*}{Diffusion transformer (DiT)}
  & Hidden dimension                & 768 \\
  & Number of layers                & 12 \\
  & Attention heads                 & 12 \\
  & MLP ratio                       & 4 \\
  & QKV bias                        & True \\
  & Register tokens                 & 24 \\
\midrule
\multirow{4}{*}{Conditioning}
  & Timestep embedding dim          & 512 \\
  & Timestep encoding               & dual (action, next-obs) \\
  & AdaLN modulation                & global obs $\oplus$ timestep \\
  & Language conditioning           & disabled \\
\midrule
\multirow{5}{*}{Diffusion process}
  & Scheduler                       & DDIM \\
  & Training timesteps              & 100 \\
  & Inference timesteps             & 10 \\
  & Noise schedule                  & cosine (\texttt{squaredcos\_cap\_v2}) \\
  & Sample clipping                 & True \\
\midrule
\multirow{5}{*}{Observation encoder}
  & Image backbone                  & ResNet-18 (ImageNet-1k) \\
  & Image resolution                & $224 \times 224$ \\
  & Color jitter / random crop      & disabled \\
  & ImageNet normalization          & disabled \\
  & Low-dim inputs                  & tool pose (7), wrench (6) \\
\midrule
\multirow{4}{*}{Future-obs latent (VAE)}
  & VAE                             & SDXL VAE (frozen) \\
  & Latent channels                 & 4 \\
  & Patch shape ($pT{,}pH{,}pW$)    & $(2, 4, 4)$ \\
  & Latent normalization (shift / scale) & $0$ / $3$ \\
\midrule
\multirow{5}{*}{Sequence \& I/O}
  & Observation history             & 4 frames \\
  & Total sequence length           & 14 frames \\
  & Action chunk length             & 7 \\
  & Action dimension                & 7 \\
  & Number of camera views          & 2 (RGB + tactile RGB) \\
\midrule
\multirow{6}{*}{Optimization}
  & Optimizer                       & AdamW \\
  & Learning rate                   & $1\!\times\!10^{-4}$ \\
  & Weight decay                    & $1\!\times\!10^{-6}$ \\
  & Adam $(\beta_1, \beta_2)$       & $(0.9,\ 0.999)$ \\
  & Gradient clip                   & 1.0 \\
  & Total training steps            & 100{,}000 \\
\bottomrule
\end{tabular}
\end{table}

\section{Physical Experiments}

\subsection{Scoring Criteria}
Both Bolt Screwing and Nut Mating require a sequence of contact-rich manipulation stages, including approach, alignment, thread engagement, rotation, and release. A binary success metric cannot fully capture partial progress, since a policy may correctly complete early stages such as alignment but fail during rotation or release. We therefore use a unified stage-based scoring protocol for both threading tasks. Each rollout is assigned a score according to the furthest valid stage reached during execution.

For both tasks, each episode starts with a human placing the manipulated object between the gripper fingers, i.e., the bolt for Bolt Screwing and the nut for Nut Mating. We evaluate 20 episodes per task with randomized initial translations, slight initial tilting of up to approximately $10^\circ$, and different initial object placements on the GelSight sensor. The threaded counterpart refers to the threaded hole in Bolt Screwing and the fixed bolt in Nut Mating. The stages are defined as follows:
\begin{itemize}
    \item \textbf{Stage 0: Failure (0 points).} The policy fails to align the manipulated object with the threaded counterpart. This includes trials where the policy stops in the air, fails to approach the target, or rotates without proper alignment.
    \item \textbf{Stage 1: Alignment (60 points).} The policy aligns the manipulated object with the threaded counterpart but does not initiate effective rotation.
    \item \textbf{Stage 2: Stuck after alignment (65 points).} The policy aligns the manipulated object with the threaded counterpart but becomes stuck before sustained rotation.
    \item \textbf{Stage 3: Rotation (75 points).} The policy maintains alignment and completes approximately a half-turn.
    \item \textbf{Stage 4: Thread engagement (95 points).} The manipulated object engages with the thread, but the gripper does not release.
    \item \textbf{Stage 5: Successful release (100 points).} The gripper releases and the manipulated object remains fixed.
\end{itemize}

In addition to the stage score, we assign a 5-point release bonus to any trial in which the policy executes a release action before achieving the final success stage. For example, if the policy reaches Stage~3 by completing approximately a half-turn and then releases the object, but the object is not fixed after release, the trial receives $75 + 5 = 80$ points. Special cases that are not explicitly covered by the main stage definitions are mapped to scores according to Table~\ref{tab:threading_special_cases}.

\begin{table}[t]
\centering
\caption{\textbf{Special-case scoring rules for the threading tasks.} These cases are used to resolve rollout behaviors that are not explicitly covered by the main stage definitions.}
\label{tab:threading_special_cases}
\begin{tabular}{p{0.68\linewidth}cc}
\toprule
\textbf{Observed behavior} & \textbf{Assigned stage} & \textbf{Score} \\
\midrule
Fails to approach the threaded counterpart. 
& Stage 0 & 0 \\
Stops in the air. 
& Stage 0 & 0 \\
Fails to align throughout the entire trial. 
& Stage 0 & 0 \\
Achieves good alignment  but fails to rotate. 
& Stage 2 & 65 \\
\bottomrule
\end{tabular}
\end{table}

\begin{table}[t]
\centering
\caption{\textbf{Stage-based grading criteria for Nut Mating and Bolt Screwing.}
Each trial is scored according to the furthest stage reached during execution.}
\label{tab:threading_grading}
\begin{tabular}{clp{0.62\linewidth}}
\toprule
\textbf{Score} & \textbf{Stage} & \textbf{Criterion} \\
\midrule
0   & Failed to align & Fails to align with the threaded counterpart. \\
60  & Aligned         & Aligns with the threaded counterpart. \\
65  & Stuck           & Aligns but becomes stuck before sustained rotation. \\
75  & Rotated $180^\circ$ & Maintains alignment and completes approximately a half-turn. \\
95 & Engaged & The manipulated object engages with the thread \\
100 & Released        & Gripper released and object is fixed.  \\
\bottomrule
\end{tabular}
\end{table}

\subsection{Detailed Results}

We evaluate each policy on the same set of 20 randomly selected initial poses for each task. Table~\ref{tab:threading_detailed_results} reports the number of trials reaching each stage, together with the total and average stage-based scores for each policy on Bolt Screwing and Nut Mating.

\begin{table}[t]
\centering
\caption{\textbf{Detailed physical experiment results.}}
\label{tab:threading_detailed_results}
\begin{tabular}{llcccccc}
\toprule
\textbf{Task} & \textbf{Method} & \textbf{Stage 0} & \textbf{Stage 2} & \textbf{Stage 3} & \textbf{Stage 5} & \textbf{Total Score} & \textbf{Avg. Score} \\
\midrule
\multirow{3}{*}{Bolt Screwing}
& MDF++ & 0 & 0 & 4 & 16 & 1900 & \textbf{95.0} \\
& UWM & 8 & 0 & 2 & 10 & 1150 & 57.5 \\
& MDF++(frozen-ViT) & 9 & 3 & 2 & 6 & 930 & 46.5 \\
\midrule
\multirow{3}{*}{Nut Mating}
& MDF++ & 1 & 1 & 10 & 8 & 1660 & \textbf{83.0} \\
& UWM & 6 & 0 & 9 & 5 & 1220 & 61.0 \\
& MDF++(frozen-ViT) & 16 & 0 & 1 & 3 & 380 & 19.0 \\
\bottomrule
\end{tabular}
\end{table}


\bibliography{example}  

@article{huang2025unified,
  title={Unified Multimodal Diffusion Forcing for Forceful Manipulation},
  author={Huang, Zixuan and Hou, Huaidian and Berenson, Dmitry},
  journal={arXiv preprint arXiv:2511.04812},
  year={2025}
}

@inproceedings{liu2024maniwav,
  title={Maniwav: Learning robot manipulation from in-the-wild audio-visual data},
  author={Liu, Zeyi and Chi, Cheng and Cousineau, Eric and Kuppuswamy, Naveen and Burchfiel, Benjamin and Song, Shuran},
  booktitle={8th Annual Conference on Robot Learning},
  year={2024}
}

@article{noseworthy2025forge,
  title={Forge: Force-guided exploration for robust contact-rich manipulation under uncertainty},
  author={Noseworthy, Michael and Tang, Bingjie and Wen, Bowen and Handa, Ankur and Kessens, Chad and Roy, Nicholas and Fox, Dieter and Ramos, Fabio and Narang, Yashraj and Akinola, Iretiayo},
  journal={IEEE Robotics and Automation Letters},
  year={2025},
  publisher={IEEE}
}

@article{kang2025robotic,
  title={Robotic compliant object prying using diffusion policy guided by vision and force observations},
  author={Kang, Jeon Ho and Joshi, Sagar and Huang, Ruopeng and Gupta, Satyandra K},
  journal={IEEE Robotics and Automation Letters},
  year={2025},
  publisher={IEEE}
}

@article{wu2024tacdiffusion,
  title={Tacdiffusion: Force-domain diffusion policy for precise tactile manipulation},
  author={Wu, Yansong and Chen, Zongxie and Wu, Fan and Chen, Lingyun and Zhang, Liding and Bing, Zhenshan and Swikir, Abdalla and Haddadin, Sami and Knoll, Alois},
  journal={arXiv preprint arXiv:2409.11047},
  year={2024}
}

@article{li2022see,
  title={See, hear, and feel: Smart sensory fusion for robotic manipulation},
  author={Li, Hao and Zhang, Yizhi and Zhu, Junzhe and Wang, Shaoxiong and Lee, Michelle A and Xu, Huazhe and Adelson, Edward and Fei-Fei, Li and Gao, Ruohan and Wu, Jiajun},
  journal={CoRL},
  year={2022}
}

@article{du2022play,
  title={Play it by ear: Learning skills amidst occlusion through audio-visual imitation learning},
  author={Du, Maximilian and Lee, Olivia Y and Nair, Suraj and Finn, Chelsea},
  journal={arXiv preprint arXiv:2205.14850},
  year={2022}
}

@article{chen2022visuo,
  title={Visuo-tactile transformers for manipulation},
  author={Chen, Yizhou and Sipos, Andrea and Van der Merwe, Mark and Fazeli, Nima},
  journal={CoRL},
  year={2022}
}

@article{zhu2025unified,
  title={Unified World Models: Coupling Video and Action Diffusion for Pretraining on Large Robotic Datasets},
  author={Zhu, Chuning and Yu, Raymond and Feng, Siyuan and Burchfiel, Benjamin and Shah, Paarth and Gupta, Abhishek},
  journal={arXiv preprint arXiv:2504.02792},
  year={2025}
}

@article{li2025unified,
  title={Unified Video Action Model},
  author={Li, Shuang and Gao, Yihuai and Sadigh, Dorsa and Song, Shuran},
  journal={arXiv preprint arXiv:2503.00200},
  year={2025}
}

@article{won2025dual,
  title={Dual-stream diffusion for world-model augmented vision-language-action model},
  author={Won, John and Lee, Kyungmin and Jang, Huiwon and Kim, Dongyoung and Shin, Jinwoo},
  journal={arXiv preprint arXiv:2510.27607},
  year={2025}
}

@inproceedings{wu2024unleashing,
  title={Unleashing large-scale video generative pre-training for visual robot manipulation},
  author={Wu, Hongtao and Jing, Ya and Cheang, Chilam and Chen, Guangzeng and Xu, Jiafeng and Li, Xinghang and Liu, Minghuan and Li, Hang and Kong, Tao},
  booktitle={International Conference on Learning Representations},
  volume={2024},
  pages={10641--10662},
  year={2024}
}

@article{hu2024video,
  title={Video prediction policy: A generalist robot policy with predictive visual representations},
  author={Hu, Yucheng and Guo, Yanjiang and Wang, Pengchao and Chen, Xiaoyu and Wang, Yen-Jen and Zhang, Jianke and Sreenath, Koushil and Lu, Chaochao and Chen, Jianyu},
  journal={arXiv preprint arXiv:2412.14803},
  year={2024}
}

@article{kim2026cosmos,
  title={Cosmos policy: Fine-tuning video models for visuomotor control and planning},
  author={Kim, Moo Jin and Gao, Yihuai and Lin, Tsung-Yi and Lin, Yen-Chen and Ge, Yunhao and Lam, Grace and Liang, Percy and Song, Shuran and Liu, Ming-Yu and Finn, Chelsea and others},
  journal={arXiv preprint arXiv:2601.16163},
  year={2026}
}

@article{liang2025video,
  title={Video generators are robot policies},
  author={Liang, Junbang and Tokmakov, Pavel and Liu, Ruoshi and Sudhakar, Sruthi and Shah, Paarth and Ambrus, Rares and Vondrick, Carl},
  journal={arXiv preprint arXiv:2508.00795},
  year={2025}
}

@article{pai2025mimic,
  title={mimic-video: Video-action models for generalizable robot control beyond vlas},
  author={Pai, Jonas and Achenbach, Liam and Montesinos, Victoriano and Forrai, Benedek and Mees, Oier and Nava, Elvis},
  journal={arXiv preprint arXiv:2512.15692},
  year={2025}
}

@article{chen2024diffusion,
  title={Diffusion forcing: Next-token prediction meets full-sequence diffusion},
  author={Chen, Boyuan and Mart{\'\i} Mons{\'o}, Diego and Du, Yilun and Simchowitz, Max and Tedrake, Russ and Sitzmann, Vincent},
  journal={Advances in Neural Information Processing Systems},
  volume={37},
  pages={24081--24125},
  year={2024}
}

@inproceedings{luo2021diffusion,
  title={Diffusion probabilistic models for 3d point cloud generation},
  author={Luo, Shitong and Hu, Wei},
  booktitle={CVPR},
  pages={2837--2845},
  year={2021}
}

@article{ho2020denoising,
  title={Denoising diffusion probabilistic models},
  author={Ho, Jonathan and Jain, Ajay and Abbeel, Pieter},
  journal={Advances in neural information processing systems},
  volume={33},
  pages={6840--6851},
  year={2020}
}

@article{guo2024prediction,
  title={Prediction with action: Visual policy learning via joint denoising process},
  author={Guo, Yanjiang and Hu, Yucheng and Zhang, Jianke and Wang, Yen-Jen and Chen, Xiaoyu and Lu, Chaochao and Chen, Jianyu},
  journal={Advances in Neural Information Processing Systems},
  volume={37},
  pages={112386--112410},
  year={2024}
}

@article{cui2024dynamo,
  title={Dynamo: In-domain dynamics pretraining for visuo-motor control},
  author={Cui, Zichen J and Pan, Hengkai and Iyer, Aadhithya and Haldar, Siddhant and Pinto, Lerrel},
  journal={Advances in Neural Information Processing Systems},
  volume={37},
  pages={33933--33961},
  year={2024}
}

@article{brandfonbrener2023inverse,
  title={Inverse dynamics pretraining learns good representations for multitask imitation},
  author={Brandfonbrener, David and Nachum, Ofir and Bruna, Joan},
  journal={Advances in Neural Information Processing Systems},
  volume={36},
  pages={66953--66978},
  year={2023}
}

@inproceedings{leng2025repa,
  title={Repa-e: Unlocking vae for end-to-end tuning of latent diffusion transformers},
  author={Leng, Xingjian and Singh, Jaskirat and Hou, Yunzhong and Xing, Zhenchang and Xie, Saining and Zheng, Liang},
  booktitle={Proceedings of the IEEE/CVF International Conference on Computer Vision},
  pages={18262--18272},
  year={2025}
}

@inproceedings{rombach2022high,
  title={High-resolution image synthesis with latent diffusion models},
  author={Rombach, Robin and Blattmann, Andreas and Lorenz, Dominik and Esser, Patrick and Ommer, Bj{\"o}rn},
  booktitle={Proceedings of the IEEE/CVF conference on computer vision and pattern recognition},
  pages={10684--10695},
  year={2022}
}

@article{luu2025manifeel,
  title={Manifeel: Benchmarking and understanding visuotactile manipulation policy learning},
  author={Luu, Quan Khanh and Zhou, Pokuang and Xu, Zhengtong and Zhang, Zhiyuan and Qiu, Qiang and She, Yu},
  journal={arXiv preprint arXiv:2505.18472},
  year={2025}
}

@article{mittal2025isaaclab,
  title={Isaac Lab: A GPU-Accelerated Simulation Framework for Multi-Modal Robot Learning},
  author={Mayank Mittal and Pascal Roth and James Tigue and Antoine Richard and Octi Zhang and Peter Du and Antonio Serrano-Muñoz and Xinjie Yao and René Zurbrügg and Nikita Rudin and Lukasz Wawrzyniak and Milad Rakhsha and Alain Denzler and Eric Heiden and Ales Borovicka and Ossama Ahmed and Iretiayo Akinola and Abrar Anwar and Mark T. Carlson and Ji Yuan Feng and Animesh Garg and Renato Gasoto and Lionel Gulich and Yijie Guo and M. Gussert and Alex Hansen and Mihir Kulkarni and Chenran Li and Wei Liu and Viktor Makoviychuk and Grzegorz Malczyk and Hammad Mazhar and Masoud Moghani and Adithyavairavan Murali and Michael Noseworthy and Alexander Poddubny and Nathan Ratliff and Welf Rehberg and Clemens Schwarke and Ritvik Singh and James Latham Smith and Bingjie Tang and Ruchik Thaker and Matthew Trepte and Karl Van Wyk and Fangzhou Yu and Alex Millane and Vikram Ramasamy and Remo Steiner and Sangeeta Subramanian and Clemens Volk and CY Chen and Neel Jawale and Ashwin Varghese Kuruttukulam and Michael A. Lin and Ajay Mandlekar and Karsten Patzwaldt and John Welsh and Huihua Zhao and Fatima Anes and Jean-Francois Lafleche and Nicolas Moënne-Loccoz and Soowan Park and Rob Stepinski and Dirk Van Gelder and Chris Amevor and Jan Carius and Jumyung Chang and Anka He Chen and Pablo de Heras Ciechomski and Gilles Daviet and Mohammad Mohajerani and Julia von Muralt and Viktor Reutskyy and Michael Sauter and Simon Schirm and Eric L. Shi and Pierre Terdiman and Kenny Vilella and Tobias Widmer and Gordon Yeoman and Tiffany Chen and Sergey Grizan and Cathy Li and Lotus Li and Connor Smith and Rafael Wiltz and Kostas Alexis and Yan Chang and David Chu and Linxi "Jim" Fan and Farbod Farshidian and Ankur Handa and Spencer Huang and Marco Hutter and Yashraj Narang and Soha Pouya and Shiwei Sheng and Yuke Zhu and Miles Macklin and Adam Moravanszky and Philipp Reist and Yunrong Guo and David Hoeller and Gavriel State},
  journal={arXiv preprint arXiv:2511.04831},
  year={2025},
  url={https://arxiv.org/abs/2511.04831}
}

@article{chi2024universal,
  title={Universal manipulation interface: In-the-wild robot teaching without in-the-wild robots},
  author={Chi, Cheng and Xu, Zhenjia and Pan, Chuer and Cousineau, Eric and Burchfiel, Benjamin and Feng, Siyuan and Tedrake, Russ and Song, Shuran},
  journal={arXiv preprint arXiv:2402.10329},
  year={2024}
}

@article{sha2026efficient,
  title={Efficient and Reliable Teleoperation through Real-to-Sim-to-Real Shared Autonomy},
  author={Sha, Shuo and Wang, Yixuan and Huang, Binghao and Loquercio, Antonio and Li, Yunzhu},
  journal={arXiv preprint arXiv:2603.17016},
  year={2026}
}

@article{zhu2026touch,
  title={Touch in the wild: Learning fine-grained manipulation with a portable visuo-tactile gripper},
  author={Zhu, Xinyue and Huang, Binghao and Li, Yunzhu},
  journal={Advances in Neural Information Processing Systems},
  volume={38},
  pages={153783--153812},
  year={2026}
}

@article{balestriero2025lejepa,
  title={Lejepa: Provable and scalable self-supervised learning without the heuristics},
  author={Balestriero, Randall and LeCun, Yann},
  journal={arXiv preprint arXiv:2511.08544},
  year={2025}
}

@article{maes2026leworldmodel,
  title={Leworldmodel: Stable end-to-end joint-embedding predictive architecture from pixels},
  author={Maes, Lucas and Lidec, Quentin Le and Scieur, Damien and LeCun, Yann and Balestriero, Randall},
  journal={arXiv preprint arXiv:2603.19312},
  year={2026}
}

@article{higuera2024sparsh,
  title={Sparsh: Self-supervised touch representations for vision-based tactile sensing},
  author={Higuera, Carolina and Sharma, Akash and Bodduluri, Chaithanya Krishna and Fan, Taosha and Lancaster, Patrick and Kalakrishnan, Mrinal and Kaess, Michael and Boots, Byron and Lambeta, Mike and Wu, Tingfan and others},
  journal={arXiv preprint arXiv:2410.24090},
  year={2024}
}

@inproceedings{podell2024sdxl,
  title={Sdxl: Improving latent diffusion models for high-resolution image synthesis},
  author={Podell, Dustin and English, Zion and Lacey, Kyle and Blattmann, Andreas and Dockhorn, Tim and M{\"u}ller, Jonas and Penna, Joe and Rombach, Robin},
  booktitle={International Conference on Learning Representations},
  volume={2024},
  pages={1862--1874},
  year={2024}
}

@article{wang2026interactive,
  title={Interactive world simulator for robot policy training and evaluation},
  author={Wang, Yixuan and Syed, Rhythm and Wu, Fangyu and Zhang, Mengchao and Onol, Aykut and Barreiros, Jose and Nayyeri, Hooshang and Dear, Tony and Zhang, Huan and Li, Yunzhu},
  journal={arXiv preprint arXiv:2603.08546},
  year={2026}
}

@article{cen2025worldvla,
  title={Worldvla: Towards autoregressive action world model},
  author={Cen, Jun and Yu, Chaohui and Yuan, Hangjie and Jiang, Yuming and Huang, Siteng and Guo, Jiayan and Li, Xin and Song, Yibing and Luo, Hao and Wang, Fan and others},
  journal={arXiv preprint arXiv:2506.21539},
  year={2025}
}

@article{zhou2024dino,
  title={Dino-wm: World models on pre-trained visual features enable zero-shot planning},
  author={Zhou, Gaoyue and Pan, Hengkai and LeCun, Yann and Pinto, Lerrel},
  journal={arXiv preprint arXiv:2411.04983},
  year={2024}
}

@inproceedings{du2024video,
  title={Video language planning},
  author={Du, Yilun and Yang, Sherry and Florence, Pete and Xia, Fei and Wahid, Ayzaan and Sermanet, Pierre and Yu, Tianhe and Abbeel, Pieter and Tenenbaum, Joshua B and Kaelbling, Leslie and others},
  booktitle={International Conference on Learning Representations},
  volume={2024},
  pages={31138--31155},
  year={2024}
}

@article{assran2025v,
  title={V-jepa 2: Self-supervised video models enable understanding, prediction and planning},
  author={Assran, Mido and Bardes, Adrien and Fan, David and Garrido, Quentin and Howes, Russell and Muckley, Matthew and Rizvi, Ammar and Roberts, Claire and Sinha, Koustuv and Zholus, Artem and others},
  journal={arXiv preprint arXiv:2506.09985},
  year={2025}
}

@article{zheng2025flare,
  title={Flare: Robot learning with implicit world modeling},
  author={Zheng, Ruijie and Wang, Jing and Reed, Scott and Bjorck, Johan and Fang, Yu and Hu, Fengyuan and Jang, Joel and Kundalia, Kaushil and Lin, Zongyu and Magne, Loic and others},
  journal={arXiv preprint arXiv:2505.15659},
  year={2025}
}

@article{team2025gigabrain,
  title={Gigabrain-0: A world model-powered vision-language-action model},
  author={Team, GigaBrain and Ye, Angen and Wang, Boyuan and Ni, Chaojun and Huang, Guan and Zhao, Guosheng and Li, Haoyun and Li, Jie and Zhu, Jiagang and Feng, Lv and others},
  journal={arXiv preprint arXiv:2510.19430},
  year={2025}
}

@article{team2025evaluating,
  title={Evaluating Gemini Robotics Policies in a Veo World Simulator},
  author={Team, Gemini Robotics and Choromanski, Krzysztof and Devin, Coline and Du, Yilun and Dwibedi, Debidatta and Gao, Ruiqi and Jindal, Abhishek and Kipf, Thomas and Kirmani, Sean and Leal, Isabel and others},
  journal={arXiv preprint arXiv:2512.10675},
  year={2025}
}

@article{zhang2025real,
  title={Real-to-sim robot policy evaluation with gaussian splatting simulation of soft-body interactions},
  author={Zhang, Kaifeng and Sha, Shuo and Jiang, Hanxiao and Loper, Matthew and Song, Hyunjong and Cai, Guangyan and Xu, Zhuo and Hu, Xiaochen and Zheng, Changxi and Li, Yunzhu},
  journal={arXiv preprint arXiv:2511.04665},
  year={2025}
}

@article{cheang2024gr,
  title={Gr-2: A generative video-language-action model with web-scale knowledge for robot manipulation},
  author={Cheang, Chi-Lam and Chen, Guangzeng and Jing, Ya and Kong, Tao and Li, Hang and Li, Yifeng and Liu, Yuxiao and Wu, Hongtao and Xu, Jiafeng and Yang, Yichu and others},
  journal={arXiv preprint arXiv:2410.06158},
  year={2024}
}

@article{oquab2023dinov2,
  title={Dinov2: Learning robust visual features without supervision},
  author={Oquab, Maxime and Darcet, Timoth{\'e}e and Moutakanni, Th{\'e}o and Vo, Huy and Szafraniec, Marc and Khalidov, Vasil and Fernandez, Pierre and Haziza, Daniel and Massa, Francisco and El-Nouby, Alaaeldin and others},
  journal={arXiv preprint arXiv:2304.07193},
  year={2023}
}

@inproceedings{assran2023self,
  title={Self-supervised learning from images with a joint-embedding predictive architecture},
  author={Assran, Mahmoud and Duval, Quentin and Misra, Ishan and Bojanowski, Piotr and Vincent, Pascal and Rabbat, Michael and LeCun, Yann and Ballas, Nicolas},
  booktitle={Proceedings of the IEEE/CVF conference on computer vision and pattern recognition},
  pages={15619--15629},
  year={2023}
}

@article{steiner2021train,
  title={How to train your vit? data, augmentation, and regularization in vision transformers},
  author={Steiner, Andreas and Kolesnikov, Alexander and Zhai, Xiaohua and Wightman, Ross and Uszkoreit, Jakob and Beyer, Lucas},
  journal={arXiv preprint arXiv:2106.10270},
  year={2021}
}

@article{zheng2026omnivta,
  title={OmniVTA: Visuo-tactile world modeling for contact-rich robotic manipulation},
  author={Zheng, Yuhang and Gu, Songen and Li, Weize and Zheng, Yupeng and Zang, Yujie and Tian, Shuai and Li, Xiang and Hao, Ce and Gao, Chen and Liu, Si and others},
  journal={arXiv preprint arXiv:2603.19201},
  year={2026}
}

@article{liu2025vitamin,
  title={Vitamin: Learning contact-rich tasks through robot-free visuo-tactile manipulation interface},
  author={Liu, Fangchen and Li, Chuanyu and Qin, Yihua and Xu, Jing and Abbeel, Pieter and Chen, Rui},
  journal={arXiv preprint arXiv:2504.06156},
  year={2025}
}

@article{huang20243d,
  title={3d-vitac: Learning fine-grained manipulation with visuo-tactile sensing},
  author={Huang, Binghao and Wang, Yixuan and Yang, Xinyi and Luo, Yiyue and Li, Yunzhu},
  journal={arXiv preprint arXiv:2410.24091},
  year={2024}
}

@inproceedings{van2016stable,
  title={Stable reinforcement learning with autoencoders for tactile and visual data},
  author={Van Hoof, Herke and Chen, Nutan and Karl, Maximilian and Van Der Smagt, Patrick and Peters, Jan},
  booktitle={2016 IEEE/RSJ international conference on intelligent robots and systems (IROS)},
  pages={3928--3934},
  year={2016},
  organization={IEEE}
}

@article{suresh2024neuralfeels,
  title={NeuralFeels with neural fields: Visuotactile perception for in-hand manipulation},
  author={Suresh, Sudharshan and Qi, Haozhi and Wu, Tingfan and Fan, Taosha and Pineda, Luis and Lambeta, Mike and Malik, Jitendra and Kalakrishnan, Mrinal and Calandra, Roberto and Kaess, Michael and others},
  journal={Science Robotics},
  volume={9},
  number={96},
  pages={eadl0628},
  year={2024},
  publisher={American Association for the Advancement of Science}
}

@inproceedings{lee2019making,
  title={Making sense of vision and touch: Self-supervised learning of multimodal representations for contact-rich tasks},
  author={Lee, Michelle A and Zhu, Yuke and Srinivasan, Krishnan and Shah, Parth and Savarese, Silvio and Fei-Fei, Li and Garg, Animesh and Bohg, Jeannette},
  booktitle={2019 International conference on robotics and automation (ICRA)},
  pages={8943--8950},
  year={2019},
  organization={IEEE}
}

@inproceedings{hansen2022visuotactile,
  title={Visuotactile-rl: Learning multimodal manipulation policies with deep reinforcement learning},
  author={Hansen, Johanna and Hogan, Francois and Rivkin, Dmitriy and Meger, David and Jenkin, Michael and Dudek, Gregory},
  booktitle={2022 International Conference on Robotics and Automation (ICRA)},
  pages={8298--8304},
  year={2022},
  organization={IEEE}
}

@inproceedings{yuan2024robot,
  title={Robot synesthesia: In-hand manipulation with visuotactile sensing},
  author={Yuan, Ying and Che, Haichuan and Qin, Yuzhe and Huang, Binghao and Yin, Zhao-Heng and Lee, Kang-Won and Wu, Yi and Lim, Soo-Chul and Wang, Xiaolong},
  booktitle={2024 IEEE International Conference on Robotics and Automation (ICRA)},
  pages={6558--6565},
  year={2024},
  organization={IEEE}
}

@inproceedings{george2025vital,
  title={Vital pretraining: Visuo-tactile pretraining for tactile and non-tactile manipulation policies},
  author={George, Abraham and Gano, Selam and Katragadda, Pranav and Farimani, Amir Barati},
  booktitle={2025 IEEE International Conference on Robotics and Automation (ICRA)},
  pages={258--264},
  year={2025},
  organization={IEEE}
}

@inproceedings{sferrazza2024power,
  title={The power of the senses: Generalizable manipulation from vision and touch through masked multimodal learning},
  author={Sferrazza, Carmelo and Seo, Younggyo and Liu, Hao and Lee, Youngwoon and Abbeel, Pieter},
  booktitle={2024 IEEE/RSJ International Conference on Intelligent Robots and Systems (IROS)},
  pages={9698--9705},
  year={2024},
  organization={IEEE}
}

@article{xu2025unit,
  title={UniT: Data efficient tactile representation with generalization to unseen objects},
  author={Xu, Zhengtong and Uppuluri, Raghava and Zhang, Xinwei and Fitch, Cael and Crandall, Philip Glen and Shou, Wan and Wang, Dongyi and She, Yu},
  journal={IEEE Robotics and Automation Letters},
  year={2025},
  publisher={IEEE}
}

@inproceedings{dave2024multimodal,
  title={Multimodal visual-tactile representation learning through self-supervised contrastive pre-training},
  author={Dave, Vedant and Lygerakis, Fotios and Rueckert, Elmar},
  booktitle={2024 IEEE International Conference on Robotics and Automation (ICRA)},
  pages={8013--8020},
  year={2024},
  organization={IEEE}
}

@article{zhao2024transferable,
  title={Transferable tactile transformers for representation learning across diverse sensors and tasks},
  author={Zhao, Jialiang and Ma, Yuxiang and Wang, Lirui and Adelson, Edward H},
  journal={arXiv preprint arXiv:2406.13640},
  year={2024}
}

@inproceedings{cao2023learn,
  title={Learn from incomplete tactile data: Tactile representation learning with masked autoencoders},
  author={Cao, Guanqun and Jiang, Jiaqi and Bollegala, Danushka and Luo, Shan},
  booktitle={2023 IEEE/RSJ International Conference on Intelligent Robots and Systems (IROS)},
  pages={10800--10805},
  year={2023},
  organization={IEEE}
}

@article{dosovitskiy2020image,
  title={An image is worth 16x16 words: Transformers for image recognition at scale},
  author={Dosovitskiy, Alexey and Beyer, Lucas and Kolesnikov, Alexander and Weissenborn, Dirk and Zhai, Xiaohua and Unterthiner, Thomas and Dehghani, Mostafa and Minderer, Matthias and Heigold, Georg and Gelly, Sylvain and others},
  journal={arXiv preprint arXiv:2010.11929},
  year={2020}
}

@article{heek2026unified,
  title={Unified Latents (UL): How to train your latents},
  author={Heek, Jonathan and Hoogeboom, Emiel and Mensink, Thomas and Salimans, Tim},
  journal={arXiv preprint arXiv:2602.17270},
  year={2026}
}
\end{document}